%% file: main.tex
\pdfoutput=1
\documentclass[letterpaper, 10 pt, conference]{ieeeconf}  %

\usepackage{amsmath,amssymb}
\usepackage{booktabs}
\usepackage[table]{xcolor}
\definecolor{fhpink}{HTML}{E75480}

\usepackage{graphicx}
\usepackage{pgf}
\usepackage{capt-of}
\let\labelindent\relax
\usepackage{enumitem}
\usepackage{microtype}
\usepackage{xspace}
\usepackage{multirow}
\makeatletter
\let\NAT@parse\@undefined
\makeatother
\usepackage[square,numbers]{natbib}
\AddToHook{env/thebibliography/begin}{\raggedright}

\makeatletter
\newcommand{\@IEEEmakecaptiontext}[2]{%
\setbox\@tempboxa\hbox{\footnotesize #1.~~ #2}%
\ifdim \wd\@tempboxa >\hsize%
\setbox\@tempboxa\hbox{\footnotesize #1.~~ }%
\parbox[t]{\hsize}{\footnotesize \noindent\unhbox\@tempboxa#2}%
\else%
\ifcenterfigcaptions \hbox to\hsize{\footnotesize\hfil\box\@tempboxa\hfil}%
\else \hbox to\hsize{\footnotesize\box\@tempboxa\hfil}%
\fi\fi}
\long\def\@makecaption#1#2{%
\ifx\@captype\@IEEEtablestring%
\@IEEEmakecaptiontext{#1}{#2}%
\@IEEEtablecaptionsepspace%
\else
\@IEEEfigurecaptionsepspace%
\@IEEEmakecaptiontext{#1}{#2}%
\fi}
\makeatother

\definecolor{linkblue}{rgb}{0,0.08,0.45}
\usepackage[colorlinks=true,linkcolor=linkblue,citecolor=linkblue,urlcolor=linkblue]{hyperref}
\hypersetup{pdfauthor={Xinchen Yu, Yoonkyo Jung, Haoran Chen, Zikui Cai, Aayush Talreja, Seungjae Lee, Hyunjin Kim, Furong Huang}, pdftitle={AuRA: From Atoms to Bits through Autonomous Robot Interaction}}

\title{\LARGE \bf
AuRA: From Atoms to Bits through Autonomous Robot Interaction
}

\author{Xinchen Yu\textsuperscript{1*}, Yoonkyo Jung\textsuperscript{1*}, Haoran Chen\textsuperscript{1}, Zikui Cai\textsuperscript{1}, Aayush Talreja\textsuperscript{1}, \\Seungjae Lee\textsuperscript{1}, Hyunjin Kim\textsuperscript{1}, and Furong Huang\textsuperscript{1,2}%
\thanks{\textsuperscript{1}University of Maryland, College Park \textsuperscript{2}All Purpose AI}
\thanks{\textsuperscript{*}Equal contribution. Email: \{xyu054, ykjung\}@umd.edu}
\thanks{Website: \href{https://aura-robot.github.io}{https://aura-robot.github.io}
}%
}

\IEEEoverridecommandlockouts
\IEEEaftertitletext{
    \begin{minipage}{\textwidth}
    \centering
    \includegraphics[width=1\textwidth]{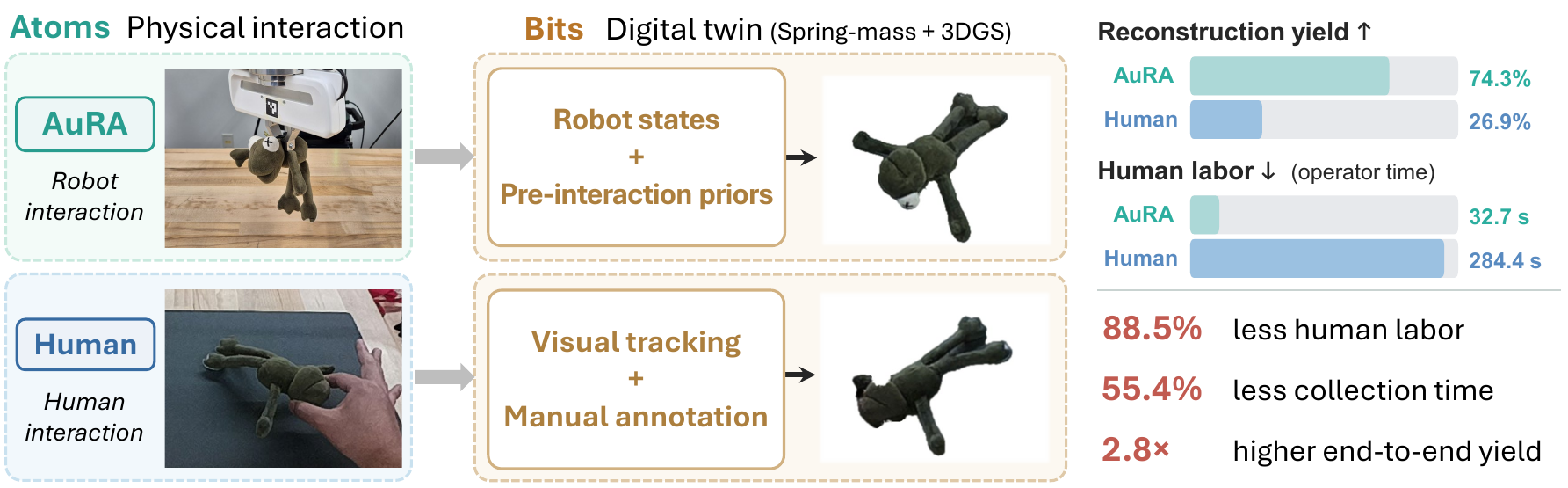}
    \vspace{-1.5em}
    \captionof{figure}{
    \textbf{AuRA: from physical interaction to digital twins.}
    AuRA replaces manual manipulation with autonomous robot interaction, derives controller motion from robot states instead of visual tracking, and uses pre-interaction priors to supplement robot-occluded regions, while retaining the same spring--mass and 3D Gaussian representation.
    Compared with human-operated collection overall, AuRA achieves 2.8$\times$ higher reconstruction yield with 88.5\% less active human labor and 55.4\% less wall-clock collection time per reconstructed asset, while maintaining broadly comparable reconstruction quality.
    }
    \vspace{1em}
    \label{fig:aura_teaser}
\end{minipage}
}

\begin{document}

\raggedbottom

\maketitle
\thispagestyle{empty}
\pagestyle{empty}

\input{sections/00_abstract}

\input{sections/01_introduction}

\input{sections/02_related}

\input{sections/03_Approach}

\input{sections/04_Experiment}

\input{sections/05_Discussion_and_Future_Work}

\input{sections/06_Acknowledgments}

\bibliographystyle{IEEEtran}
\bibliography{example}

\end{document}

%% file: sections/00_abstract.tex
\begin{abstract}
Simulation enables robot-learning experiments at scale, but acquiring physics-informed digital twins often requires human manipulation and manual annotation of recorded interactions. To reduce this reliance on human effort, we present AuRA, a system that automates robot interaction and data preparation for digital-twin reconstruction. AuRA combines execution-aware grasp selection, gripper-motion measurements derived from robot states, and registered pre-interaction geometry and appearance priors. These choices enable autonomous collection, avoid reliance on visual gripper tracking, and recover object information obscured by the robot. Across 13 deformable objects, AuRA reduces active human labor and wall-clock collection time per successfully reconstructed asset by 88.5\% and 55.4\%, respectively. Relative to the human-collected PhysTwin baseline, AuRA improves seven of nine metrics for future prediction under known gripper motion, including a 28.1\% reduction in one-way $L_1$ Chamfer distance. It also achieves similar mean silhouette IoU and Chamfer distance under the evaluated unseen interactions. These results highlight the potential of autonomous robot interaction to support more efficient digital-twin acquisition with comparable reconstruction quality.
\end{abstract}

%% file: sections/01_introduction.tex
\section{Introduction}

Simulation enables repeated robot-learning experiments in digital environments, supporting scalable data generation and policy training~\citep{gong2023arnold,margolisyang2022rapid,Acar_2024}. To reflect interactions with particular real-world objects, these environments need digital assets that represent the objects' geometry, appearance, and physical behavior. Real-to-sim reconstruction offers a way to build such assets from observations of real objects, with manipulation providing information about their dynamics. The resulting assets can be reused across simulated experiments, supporting a shift from an ``atoms loop'' of physical manipulation to a ``bits loop'' of simulation.

A key bottleneck, however, lies in acquiring the real-world observations required for reconstruction. Existing workflows often rely on human manipulation or manual annotation of recorded data~\citep{torne2024rialto,phystwin}. This per-object human involvement limits the rate at which asset collections can be expanded. Automating interaction-data acquisition offers a way to reduce this effort: a robot can select and execute interactions while recording the observations needed for reconstruction. In practice, a reliable autonomous pipeline must address three coupled challenges.

\emph{First, selected interactions must satisfy robot and workspace constraints.} A visually plausible grasp may violate joint limits or bring the gripper too close to the table or cameras, creating collision risks.
\emph{Second, reconstruction requires reliable gripper-motion inputs.} Segmentation, tracking, and depth errors can corrupt visually estimated gripper trajectories, potentially biasing the physical parameters inferred from the object's response.
\emph{Third, robot-induced occlusion can leave object geometry and appearance incomplete.} The gripper and arm may obscure portions of the object throughout the trajectory. Missing surface observations can leave gaps in both the physical model and its rendered appearance.

To address these challenges, we introduce \textbf{AuRA} (\textbf{Au}tonomous \textbf{R}eal-to-sim \textbf{A}sset Reconstruction), a system for autonomously collecting object-interaction data and reconstructing simulation-ready assets. First, AuRA autonomously identifies and executes feasible object interactions. After object placement and system initialization, AuRA generates grasp candidates from pre-interaction RGB-D observations and checks their feasibility against robot and workspace constraints. The robot then executes the selected interaction while the system records RGB-D videos and robot states.

Second, AuRA separates measurement of the robot's action from observation of the object's response. Recorded robot states and calibrated gripper geometry are used to derive controller trajectories, while visual tracking captures the object's motion and deformation. This separation provides reliable motion inputs without requiring visual gripper segmentation, tracking, or depth lifting.

Third, AuRA reuses pre-interaction observations to mitigate robot-induced occlusion. It constructs a static point cloud and a Gaussian appearance model, then aligns both with the first interaction frame to account for small pose changes caused by grasping. These registered priors supplement geometry and appearance missing from the interaction observations. The same pre-interaction captures therefore support both physical acquisition and digital reconstruction, without requiring a prebuilt object asset or a separate manual capture stage.

We evaluate data collection efficiency, reconstruction reliability, and asset quality across 13 deformable objects. Compared with human-operated collection overall, AuRA reduces active human labor and wall-clock collection time per successfully reconstructed asset by 88.5\% and 55.4\%, respectively. Across all acquisition attempts in the timed collection study, AuRA achieves an end-to-end reconstruction success rate of 74.3\%, compared with 26.9\% for human-operated collection overall. Moreover, AuRA improves seven of nine metrics compared to the human-collected PhysTwin baseline for future prediction, reducing one-way $L_1$ Chamfer distance by 28.1\% and increasing depth coverage by 9.9 percentage points. The two methods also achieve similar mean silhouette IoU and Chamfer distance on the evaluated unseen interactions. In summary, our main contributions are:
\begin{enumerate}[leftmargin=*]
    \item \textbf{Autonomous data acquisition for digital-twin reconstruction.} We introduce AuRA, a system that integrates execution-aware grasp selection, autonomous object interaction, and synchronized recording of RGB-D observations and robot states. After setup and initialization, collection proceeds without manual manipulation or annotation.
    \item \textbf{Reliable controller trajectories from robot states.} We derive controller-point trajectories from recorded robot joint states and calibrated gripper geometry. Together with visually tracked object motion, these trajectories support dynamics reconstruction without relying on visual gripper segmentation, tracking, or depth lifting.
    \item \textbf{Reusing pre-interaction observations for asset reconstruction.} We align pre-interaction geometry priors with the recorded interaction to fill gaps caused by robot occlusion, providing a more complete geometric initialization for dynamics optimization. Together with 3D Gaussians trained from the same pre-interaction captures, AuRA maintains reconstruction quality while significantly reducing human labor through autonomous data collection.
\end{enumerate}

%% file: sections/02_related.tex
\section{Related Work}

\subsection{Reconstruction and Simulation of Deformable Objects}

Geometric and visual reconstruction methods recover object shape and appearance using representations such as point clouds, neural radiance fields, and 3D Gaussians~\citep{wang2025vggt,mildenhall2020nerf,kerbl3Dgaussians}. Digital twins for deformable-object interaction additionally require dynamics that predict how shape changes under applied motion or forces.

One line of work uses neural networks to learn predictive models of object dynamics from data. Graph-based simulators represent objects as particles or nodes and learn to predict their motion~\citep{shi2022robocraft,sanchezgonzalez2020learning,lin2022learning}. Recent systems further couple learned dynamics with Gaussian appearance representations to support both motion prediction and rendering~\citep{zhang2024particle,huang2026soma}.

An alternative fits explicit physical models to observed behavior, using differentiable mesh simulation~\citep{Qiao2021Differentiable}, spring--mass systems~\citep{phystwin,zhong2024springgaus,lee2026physhandi}, or the material point method~\citep{MPM,chenhu2026empm}. Within this family, PhysTwin~\citep{phystwin} reconstructs geometry, appearance, and dynamics from sparse-view RGB-D videos; PhysHanDI~\citep{lee2026physhandi} emphasizes coupled hand--object reconstruction; and EMPM~\citep{chenhu2026empm} combines offline identification with online material adaptation. Their demonstrated acquisition pipelines rely on human manipulation or teleoperation. AuRA builds on this reconstruction paradigm and addresses the complementary challenge of autonomously acquiring the interaction data needed for reconstruction.

\subsection{Robotic Interaction for Object Modeling}

Beyond reconstruction from recorded observations, another line of work uses robotic interaction to actively acquire information for object modeling. Autonomous soft-object manipulation~\citep{hu2026autonomous}, viewpoint and object-reorientation selection~\citep{Dasgupta2024ActNeRF}, and active visuotactile cable exploration~\citep{mazza2026activecrossmodalvisuotactileperception} improve observations of object shape or appearance. These methods focus on geometry and/or appearance reconstruction, without jointly modeling deformation dynamics.

Other systems use interaction to identify physical properties or adapt predictive models. DPSI~\citep{Yang2025DPSI} estimates material and environment parameters through differentiable simulation, while AdaptiGraph adapts learned dynamics through active interaction~\citep{zhang2024adaptigraph}; neither jointly reconstructs a renderable appearance model. Sim2Real$^2$~\citep{Ma2023Sim2Real2} and SAGCI-System~\citep{lv2022sagci} construct or refine articulated-object models for planning and control, using rigid-link representations rather than continuous deformation within links. AuRA brings autonomous data acquisition to digital-twin reconstruction encompassing geometry, appearance, and deformation dynamics.

%% file: sections/03_Approach.tex
\section{AuRA: Autonomous Data Preparation and Reconstruction}
\label{sec:method}

\begin{figure*}[t!]
    \centering
    \includegraphics[width=1\textwidth]{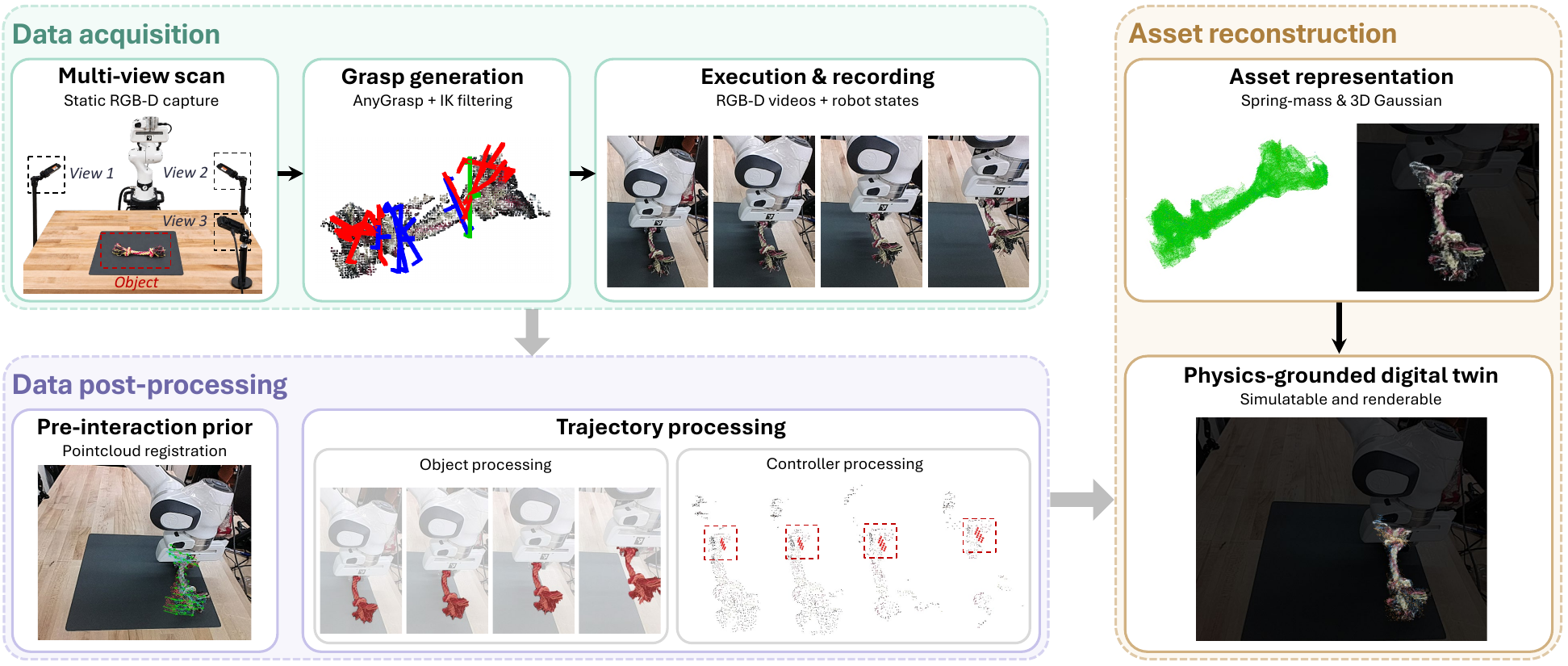}
    \caption{
        \textbf{AuRA's autonomous physical-to-digital acquisition pipeline.} Multi-view RGB-D capture supports grasp generation and execution-feasibility filtering. The robot executes the selected grasp and records the interaction with synchronized robot states. During processing, visual tracking provides object trajectories, while robot poses and calibrated gripper geometry provide controller trajectories. AuRA also constructs a static point cloud and trains a Gaussian appearance model from its pre-interaction captures, then aligns both with the first interaction frame to supplement robot-occluded regions. These inputs support spring--mass optimization and Gaussian rendering through the PhysTwin reconstruction backend.
    }
    \label{fig:robot_driven_phystwin_pipeline}
\end{figure*}

To address the challenges of autonomous acquisition, we propose AuRA, a pipeline with three stages: (1) autonomous data acquisition, (2) data post-processing, and (3) physics-informed asset reconstruction, as shown in Fig~\ref{fig:robot_driven_phystwin_pipeline}. The first stage uses pre-interaction RGB-D observations to generate and filter grasp candidates, then executes a selected grasp-and-lift interaction while recording synchronized RGB-D videos and robot joint states. The second stage extracts object and controller trajectories and registers the pre-interaction static point cloud to supplement missing object geometry, preparing the inputs for physical reconstruction. The third stage fits a spring--mass model using these inputs and trains a Gaussian appearance model using the pre-interaction observations. The Gaussians are then aligned and bound to the physical model, yielding a digital twin that represents geometry, appearance, and dynamics applicable to both rigid and deformable objects.

\subsection{Data Acquisition}
\label{sec:method_collection}

After object placement and system initialization, AuRA acquires interaction data without manual object manipulation or recording-boundary annotation. Before manipulation, calibrated RGB-D cameras capture the stationary object from multiple viewpoints. AuRA segments the object in each view using a text prompt derived from a user-provided case name, then back-projects valid depth measurements within the masks and fuses them into a point cloud in the common world frame. The cloud is filtered to suppress spurious points caused by depth noise and RGB--depth misalignment, using the parameters in Table~\ref{tab:grasp_pointcloud_filtering}.

\begin{table}[t]
    \centering
    \caption{\textbf{Point-cloud preprocessing parameters for grasp pose generation.}}
    \label{tab:grasp_pointcloud_filtering}
    \small
    \begin{tabular}{ll}
        \hline
        Parameter & Default value \\
        \hline
        Valid depth measurement range & $0.2 < d < 1.5$ m \\
        Reference table height $z_{\mathrm{table}}$ & $0$ m \\
        Below-table tolerance $\epsilon$ & $0.01$ m \\
        Downsampling voxel size & $0.003$ m \\
        Neighbors for statistical filtering & $35$ \\
        Statistical rejection threshold & $1.2$ standard deviations \\
        Local neighborhood radius & $0.006$ m \\
        Required neighbors within radius & $10$ \\
        Principal-axis cropping percentiles & $1$st--$99$th \\
        Cropping margin along each axis & $0$ m \\
        \hline
    \end{tabular}
\end{table}

We then use AnyGrasp~\citep{fang2023anygrasp} to generate candidate grasp poses and gripper opening widths from the filtered point cloud. AuRA checks tabletop clearance using the end-effector center and calibrated fingertip positions. It also uses PyRoki~\citep{kim2025pyroki} to solve inverse kinematics and check joint-limit margins at sampled poses along the planned motion path. The robot then performs the highest-confidence candidate that passes these checks while AuRA records synchronized RGB-D videos and robot joint states. Recording stops automatically when the lift is complete.

\input{sections/collection_efficiency_figure}

\subsection{Data Post-processing}
\label{sec:method_reconstruction}

After data acquisition, AuRA prepares object trajectories, controller trajectories, and a supplemented object point cloud for reconstruction. It segments the object in the interaction videos using Grounded-SAM2~\citep{ren2024grounded} and initializes tracking points from the filtered first-frame point cloud. MV-TAP~\citep{koo2025mvtap} jointly tracks points initially visible in multiple cameras, while CoTracker3~\citep{karaev2024cotracker3} handles those initially visible in only one camera. The resulting 2D tracks are lifted into 3D using the corresponding depth maps and camera calibration, then filtered for visibility, valid depth, and local motion consistency.

Controller trajectories are derived deterministically from recorded robot joint states. The resulting end-effector poses, calibrated fingertip geometry, and robot-to-world transformation determine the paths of gripper-attached controller points that drive the simulated object. This avoids visual gripper segmentation, tracking, and depth lifting, providing controller-motion inputs even when the gripper is occluded.

To supplement geometry obscured by the robot, AuRA registers the pre-interaction static point cloud to the visible object geometry in the first interaction frame. Assuming that grasping causes only small shape changes before recording begins, registration uses a rigid transformation estimated through PCA initialization and ICP refinement. An alignment is accepted only if it passes geometric checks and sufficiently reduces the trimmed symmetric Chamfer distance; otherwise, the identity transformation is retained. The registered cloud adds surface points only in voxels unoccupied by tracked object points, preserving the tracked observations while filling missing regions. This supplemented geometry provides a more complete geometric initialization for dynamics optimization.

\subsection{Asset Reconstruction}
\label{sec:physical_visual_representation}

Using the supplemented object geometry and the object and controller trajectories, AuRA fits a physical model through PhysTwin's spring--mass optimization~\citep{phystwin}. Gripper-attached controller points drive the simulated object through springs connecting controller and object nodes, while the tracked object trajectories supervise the optimization.

For appearance, AuRA trains a static 3D Gaussian model from the pre-interaction RGB-D observations, initialized with the static point cloud. These observations provide appearance information for regions later obscured by the robot. The rigid transformation obtained during point-cloud registration is applied to the Gaussian positions and orientations to align the appearance model with the interaction sequence.

The aligned Gaussians are then bound to the fitted physical model through weighted blending of nearby mass-node motion. This binding transfers simulated deformation to the Gaussians for rendering, yielding a digital twin that represents object geometry, appearance, and dynamics.

%% file: sections/collection_efficiency_figure.tex
\begin{figure*}[t]
    \centering
    \includegraphics[width=1\textwidth]{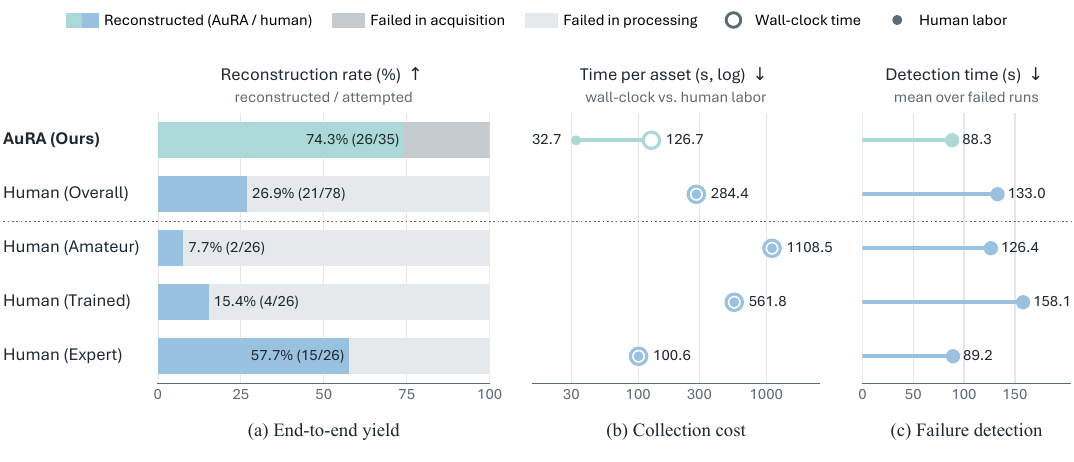}
    \vspace{-2em}
    \caption{\textbf{Data collection efficiency and reconstruction reliability.} Human (Overall) pools the three groups below the dotted line. (a)~End-to-end reconstruction rate: AuRA fails during acquisition, whereas human recordings fail during data processing. (b)~Wall-clock time ($\circ$) and human labor ($\bullet$) per reconstructed asset (log scale): the two are equal for human collection. (c)~Mean failure detection time: AuRA detects failures after 88.3~s on average, compared with 133.0~s for human collection overall and 89.2~s for experts. Collection costs in (b) exclude data processing, reconstruction training, and untimed backup trials.}
    \label{fig:collection_efficiency}
\end{figure*}

%% file: sections/04_Experiment.tex
\section{Experiment}

Our setup consists of a gripper-equipped Franka robot arm and three OAK-D RGB-D cameras positioned around a tabletop workspace. The cameras and robot are calibrated to a common world frame.

We evaluate data collection efficiency, reliability, and reconstruction quality on 13 deformable objects. Each trial begins with the object placed within the workspace and consists of a single grasp-and-lift interaction for either human-operated or robot-driven collection. We use the human-collected PhysTwin pipeline~\citep{phystwin} as our baseline because it reconstructs geometry, appearance, and dynamics from limited interaction data. The baseline uses CoTracker3~\citep{karaev2024cotracker3} for object tracking. The default AuRA pipeline uses MV-TAP~\citep{koo2025mvtap} for object tracking with a CoTracker3 fallback for points initially visible in only one camera, the registered pre-interaction static point cloud as a geometry prior, and static 3D Gaussians for appearance. Data collection and PhysTwin baseline training run on a desktop equipped with an NVIDIA RTX 5090 GPU, while AuRA reconstruction runs on a server equipped with an NVIDIA RTX A5000 GPU.

\begin{table*}[t]
    \centering

    \caption{\textbf{Resimulation and future-prediction quality.}
    Trajectory-level metrics are averaged within each object and then equally across 13 objects.
    PhysTwin uses 29 human trajectories, with shape priors when available; AuRA uses 26 robot
    trajectories processed by its standard pipeline. PhysTwin (CoTracker3) and AuRA (MV-TAP)
    retain an average of 1,493.76 and 615.96 tracking points per trajectory, respectively.
    All available views with usable object masks are used in evaluation.}

    \label{tab:reconstruction_quality}

    \small
    \setlength{\tabcolsep}{6pt}
    \renewcommand{\arraystretch}{1.03}

    \begin{tabular*}{0.78\textwidth}{
        @{\extracolsep{\fill}}
        l
        c c
        @{\hspace{10pt}}
        c c
        @{}
    }
        \toprule

        \multirow{2}{*}{\textbf{Metric}}
        & \multicolumn{2}{c}{\textbf{Resimulation}}
        & \multicolumn{2}{c}{\textbf{Future Prediction}} \\

        \cmidrule(lr){2-3}
        \cmidrule(lr){4-5}

        &
        \textbf{PhysTwin}
        & \textbf{AuRA}
        & \textbf{PhysTwin}
        & \textbf{AuRA} \\

        \midrule

        PSNR (dB) $\uparrow$
        & \textbf{12.289}
        & 11.549
        & 10.325
        & \textbf{10.590} \\

        SSIM $\uparrow$
        & \textbf{0.9856}
        & 0.9852
        & \textbf{0.9805}
        & 0.9797 \\

        LPIPS $\downarrow$
        & \textbf{0.0126}
        & 0.0148
        & 0.0233
        & \textbf{0.0230} \\

        IoU $\uparrow$
        & \textbf{0.6675}
        & 0.6090
        & 0.4607
        & \textbf{0.5001} \\

        Precision $\uparrow$
        & \textbf{0.7521}
        & 0.7021
        & 0.5899
        & \textbf{0.6132} \\

        Recall $\uparrow$
        & \textbf{0.7972}
        & 0.7791
        & 0.6260
        & \textbf{0.7085} \\

        Depth MAE (mm) $\downarrow$
        & \textbf{21.797}
        & 24.093
        & \textbf{27.784}
        & 34.923 \\

        Depth Coverage $\uparrow$
        & \textbf{0.8152}
        & 0.8050
        & 0.6392
        & \textbf{0.7386} \\

        One-Way $L_1$ Chamfer (mm) $\downarrow$
        & 11.225
        & \textbf{10.104}
        & 23.335
        & \textbf{16.783} \\

        \bottomrule
    \end{tabular*}

\end{table*}

\begin{table*}[t]
    \centering
    \caption{\textbf{First-frame alignment and unseen-interaction quality.}
    The upper block reports silhouette metrics at the aligned first frame; the lower block evaluates the subsequent interaction.
    Bold indicates the best value in each row.}
    \label{tab:unseen_interaction_quality}

    \small
    \setlength{\tabcolsep}{8pt}
    \renewcommand{\arraystretch}{1.12}

    \begin{tabular}{l ccc}
        \toprule

        \multirow{2}{*}{\textbf{Metric}}
        & \multirow{2}{*}{\textbf{PhysTwin}}
        & \multicolumn{2}{c}{\textbf{AuRA}} \\

        \cmidrule(lr){3-4}

        &
        & \textbf{Static GS}
        & \textbf{First-Frame GS} \\

        \midrule

        \multicolumn{4}{l}{\textit{First-Frame Alignment}} \\
        \addlinespace[2pt]

        Mask IoU $\uparrow$
        & \textbf{0.5096}
        & 0.4939
        & 0.4095 \\

        Mask Precision $\uparrow$
        & \textbf{0.6228}
        & 0.5817
        & 0.5285 \\

        Mask Recall $\uparrow$
        & 0.7146
        & \textbf{0.7650}
        & 0.6752 \\

        \midrule

        \multicolumn{4}{l}{\textit{Unseen Interaction}} \\
        \addlinespace[2pt]

        PSNR (dB) $\uparrow$
        & 10.309
        & \textbf{10.379}
        & 9.622 \\

        SSIM $\uparrow$
        & \textbf{0.9842}
        & 0.9815
        & 0.9798 \\

        LPIPS $\downarrow$
        & \textbf{0.0233}
        & 0.0256
        & 0.0288 \\

        \addlinespace[6pt]

        Mask IoU $\uparrow$
        & 0.3849
        & \textbf{0.3861}
        & 0.3294 \\

        Mask Precision $\uparrow$
        & \textbf{0.4673}
        & 0.4444
        & 0.4020 \\

        Mask Recall $\uparrow$
        & 0.5991
        & \textbf{0.6522}
        & 0.5821 \\

        \addlinespace[6pt]

        Depth MAE (mm) $\downarrow$
        & 18.936
        & 19.522
        & \textbf{18.478} \\

        Depth Coverage $\uparrow$
        & 0.5991
        & \textbf{0.6522}
        & 0.5821 \\

        One-Way $L_1$ Chamfer (mm) $\downarrow$
        & \textbf{43.430}
        & 43.554
        & 43.554 \\

        \bottomrule
    \end{tabular}
\end{table*}

\subsection{Data Collection Efficiency and Reliability}

We compare the collection cost and reliability of AuRA with human-operated acquisition across three experience groups, namely amateur, trained, and expert. Each human group contributes two trajectories per object. Eight additional untimed trials supplement the reconstruction-quality evaluation to ensure at least two successfully reconstructed assets per object; they are excluded from the efficiency comparison.

As shown in Fig.~\ref{fig:collection_efficiency}, AuRA achieves both a shorter mean failure-detection time and a higher end-to-end yield than all human groups. AuRA detects failed runs after an average of 88.3s and achieves a yield of 74.3\%, compared with 126.4s and 7.7\% for the amateur group, 158.1s and 15.4\% for the trained group, and 89.2s and 57.7\% for the expert group. Human collection overall requires 133.0s to detect failures and achieves a yield of 26.9\%. Specifically, AuRA produces 26 reconstructed assets from 35 acquisition attempts, whereas human-operated collection produces 21 from 78 attempts overall. We attribute AuRA's higher reliability to two design choices: gripper-pose feasibility filtering rejects infeasible interaction candidates before execution, while controller trajectories derived deterministically from robot states avoid downstream reconstruction failures caused by visual gripper tracking, as further examined in the ablation studies.

For efficiency, we measure wall-clock collection time and active human labor per successfully reconstructed asset. Costs are computed over all timed acquisition attempts, including failed attempts, while excluding downstream data processing and reconstruction, including PhysTwin's failure-detection processing. Again, as demonstrated in Fig.~\ref{fig:collection_efficiency}, AuRA requires 126.7s of wall-clock collection time and 32.7s of active human labor per asset, corresponding to reductions of 55.4\% and 88.5\%, respectively, relative to human collection overall, which requires 284.4s for both measures. AuRA also incurs substantially less human labor than each individual human group, compared with 1,108.5s for amateurs, 561.8s for trained users, and 100.6s for experts. Its wall-clock collection time is lower than that of the amateur and trained groups, though higher than the 100.6s required by experts. Because AuRA requires human involvement only for object placement and execution initialization, it eliminates the manual effort associated with performing interactions and annotating recording boundaries. Combined with its high reconstruction success rate, this reduction in manual effort enables more efficient acquisition of reconstructed assets.

\begin{figure*}[t]
    \centering
    \begingroup
    \newcommand{\unseenimage}[2]{\includegraphics[width=\linewidth,trim={#1},clip]{images/#2}}
    \setlength{\tabcolsep}{2pt}
    \renewcommand{\arraystretch}{1.1}
    \begin{tabular}{@{}p{0.10\textwidth}p{0.214\textwidth}p{0.214\textwidth}p{0.214\textwidth}p{0.214\textwidth}@{}}
        & \centering\footnotesize\shortstack{Data collection\\frame 0 (GT)} & \centering\footnotesize\shortstack{Rendering aligned to\\unseen interaction frame 0} & \centering\footnotesize\shortstack{Rendering overlay} & \multicolumn{1}{c}{\footnotesize\shortstack{Unseen interaction\\frame 0 (GT)}} \\
        \raisebox{35pt}[0pt][0pt]{\parbox[c]{\linewidth}{\footnotesize AuRA\\Static GS}}
        & \unseenimage{560bp 170bp 220bp 220bp}{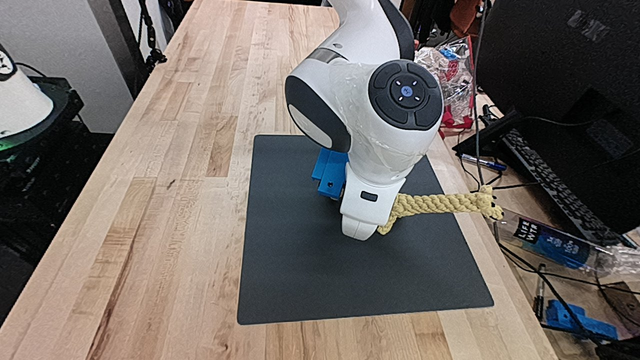}
        & \unseenimage{134.4bp 40.8bp 52.8bp 52.8bp}{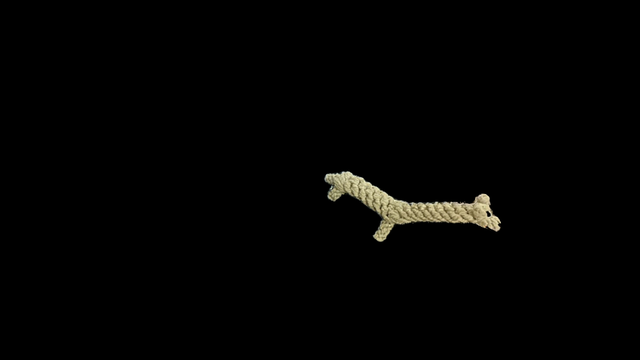}
        & \unseenimage{134.4bp 40.8bp 52.8bp 52.8bp}{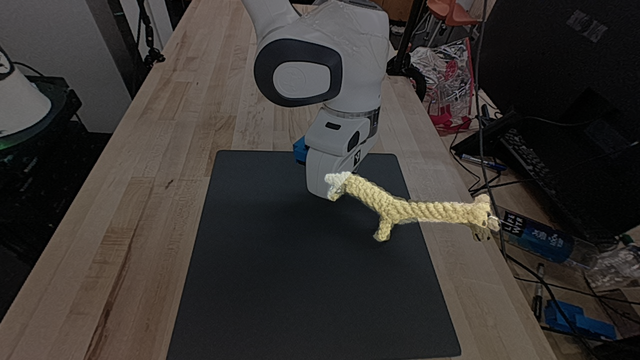}
        & \unseenimage{560bp 170bp 220bp 220bp}{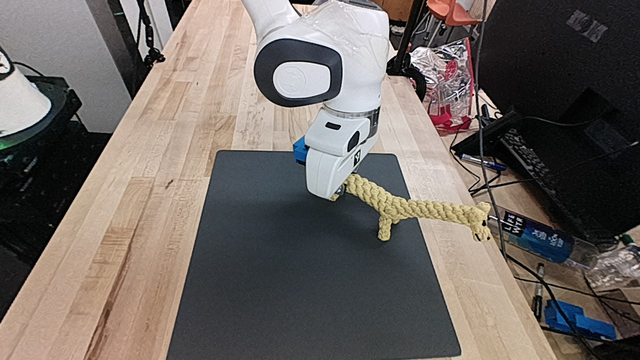} \\
        \raisebox{35pt}[0pt][0pt]{\parbox[c]{\linewidth}{\footnotesize AuRA\\First-frame GS}}
        & \unseenimage{560bp 170bp 220bp 220bp}{AuRA_reconstruction_data_frame0.png}
        & \unseenimage{134.4bp 40.8bp 52.8bp 52.8bp}{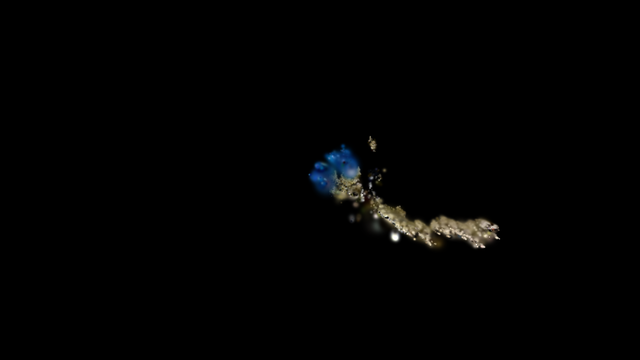}
        & \unseenimage{134.4bp 40.8bp 52.8bp 52.8bp}{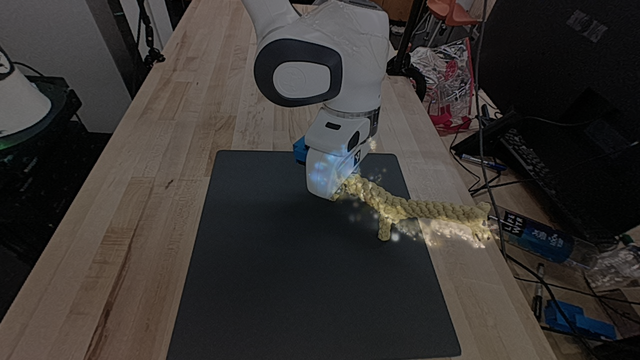}
        & \unseenimage{560bp 170bp 220bp 220bp}{AuRA_gt.png} \\
        \raisebox{35pt}[0pt][0pt]{\parbox[c]{\linewidth}{\footnotesize PhysTwin}}
        & \unseenimage{540bp 145bp 240bp 245bp}{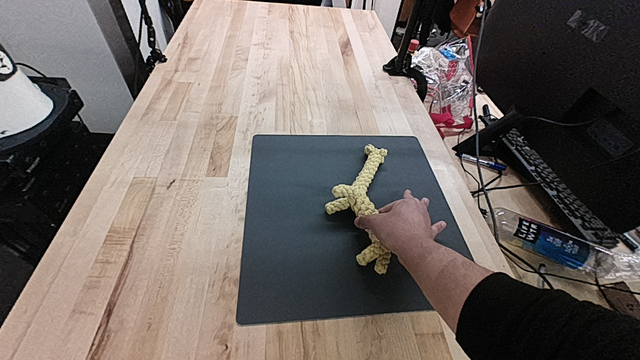}
        & \unseenimage{129.6bp 34.8bp 57.6bp 58.8bp}{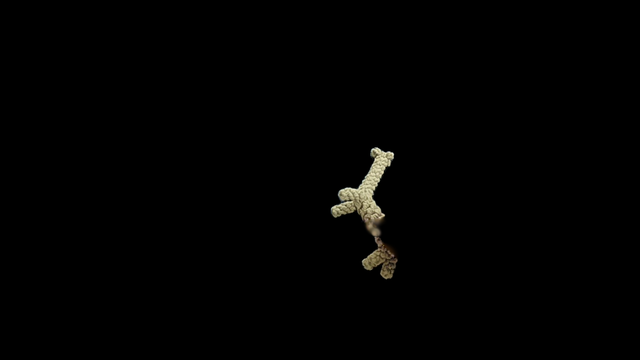}
        & \unseenimage{129.6bp 34.8bp 57.6bp 58.8bp}{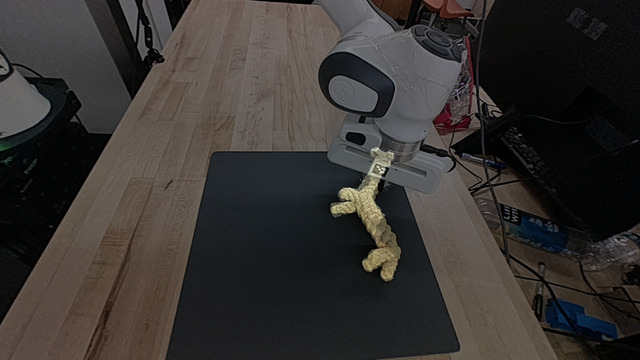}
        & \unseenimage{540bp 145bp 240bp 245bp}{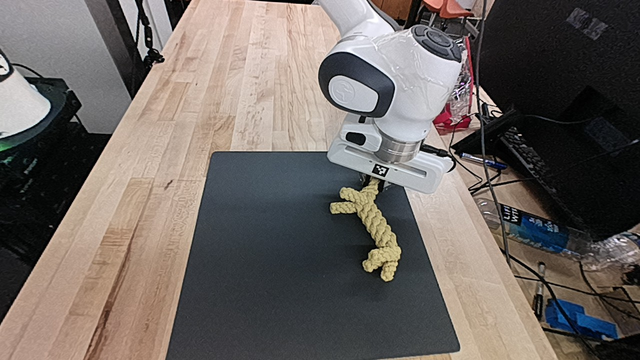} \\
    \end{tabular}
    \endgroup
    \caption{\textbf{Object appearance at the first frame of an unseen interaction.} From left to right: the original data-collection frame 0, the rendered object at unseen interaction frame 0 after manual asset alignment, its overlay with the unseen-interaction ground truth (GT), and that GT image.}
    \label{fig:unseen_appearance_comparison}
\end{figure*}

\subsection{Reconstruction Quality}
\subsubsection{Resimulation and Future Prediction}

Following PhysTwin~\citep{phystwin}, we use the first 70\% of the frames in each interaction sequence for training and the remaining 30\% for testing. The learned spring--mass model is driven by recorded controller trajectories, and its motion animates the associated 3D Gaussians for rendering. Resimulation measures agreement with observations over the training frames, while future prediction evaluates the held-out frames under known controller motion. The mean sequence lengths are 34.62 frames for the human baseline and 33.31 frames for AuRA.

As shown in Table~\ref{tab:reconstruction_quality}, AuRA outperforms PhysTwin on seven of nine future-prediction metrics, whereas PhysTwin performs better on eight of nine resimulation metrics. In future prediction, AuRA shows particularly clear gains in silhouette and geometry, improving mask IoU from 0.4607 to 0.5001, recall from 0.6260 to 0.7085, and depth coverage from 0.6392 to 0.7386, while reducing the one-way $L_1$ Chamfer distance from 23.335~mm to 16.783~mm. AuRA also achieves slightly better PSNR and LPIPS, although PhysTwin retains higher SSIM and lower depth MAE. In contrast, PhysTwin performs consistently better during resimulation, with AuRA improving only the one-way $L_1$ Chamfer distance. One possible explanation is that PhysTwin retains substantially more tracking points, providing denser supervision for fitting the observed motion. We further examine this effect through tracker and initialization ablations in Table~\ref{tab:tracker_ablation} and discuss the retained tracking-point counts in ablation studies.

\subsubsection{Unseen Interactions}
\label{sec:unseen_interactions}

For each of the 13 objects, we select one asset from each method and collect one new interaction per asset, resulting in 26 unseen-interaction trajectories. Assets are selected based on their original future-prediction performance; for AuRA, we prioritize cases with both static and first-frame Gaussian models, while shape-prior-enabled PhysTwin assets are used when available. Before each interaction, the physical object is manually aligned with the selected asset, and the Franka robot performs a teleoperated drag-and-lift interaction. Recorded robot states drive the reconstructed dynamics, and the resulting sequence is evaluated against the recorded observations using the same metrics as above.

As shown in Table~\ref{tab:unseen_interaction_quality}, AuRA with static Gaussians and PhysTwin achieve similar mean performance on the unseen interactions. AuRA attains a mask IoU of 0.3861, compared with 0.3849 for PhysTwin, while their one-way $L_1$ Chamfer distances are 43.554~mm and 43.43~mm, respectively. Appearance quality is also similar: AuRA achieves 10.379~dB PSNR, 0.9815 SSIM, and 0.0256 LPIPS, compared with 10.309~dB, 0.9842, and 0.0233 for PhysTwin.

The comparison between AuRA's static and first-frame Gaussian models further highlights the benefit of pre-interaction priors. Static Gaussians improve all reported appearance and silhouette metrics over first-frame Gaussians, preserving visual information from regions that become occluded by the robot during interaction, as illustrated in Fig.~\ref{fig:unseen_appearance_comparison}. The registered static point cloud provides a complementary geometry prior by supplying surface observations missing from the interaction sequence. Together, these priors reduce the impact of robot-induced occlusion on appearance and dynamics reconstruction.

\subsection{Ablation Studies}

\begin{table*}[t]
    \centering

    \caption{\textbf{Object-tracker and pre-interaction static geometry-prior ablations.}
    All configurations use the same 26 robot-collected trajectories and static Gaussians.
    Each cell reports results without / with the geometry prior.
    CoTracker3-m refers to mask-based initialization and CoTracker3-p refers to point-cloud-based initialization, respectively.
    Bold indicates the best score across all methods within each prior setting, and
    underlining marks the better CoTracker3 initialization. Both use unrounded values.}

    \label{tab:tracker_ablation}

    \small
    \setlength{\tabcolsep}{3.5pt}
    \renewcommand{\arraystretch}{1.03}

    \begin{tabular*}{\textwidth}{
        @{\extracolsep{\fill}}
        l
        cccc
        @{}
    }
        \toprule

        \textbf{Metric}
        & \textbf{CoTracker3-m}
        & \textbf{CoTracker3-p}
        & \textbf{MVTracker}
        & \textbf{MV-TAP} \\

        \midrule

        \multicolumn{5}{@{}l}{\textit{Resimulation}} \\
        \addlinespace[2pt]

        PSNR (dB) $\uparrow$
        & \underline{11.341} / \underline{\textbf{11.629}}
        & 11.245 / 11.520
        & 11.445 / 11.543
        & \textbf{11.468} / 11.549 \\

        SSIM $\uparrow$
        & 0.9843 / \underline{\textbf{0.9852}}
        & \underline{0.9843} / 0.9851
        & 0.9847 / 0.9851
        & \textbf{0.9848} / 0.9852 \\

        LPIPS $\downarrow$
        & 0.0173 / \underline{\textbf{0.0148}}
        & \underline{0.0172} / 0.0149
        & 0.0162 / 0.0149
        & \textbf{0.0159} / 0.0148 \\

        \addlinespace[4pt]

        Mask IoU $\uparrow$
        & \underline{0.5651} / \underline{\textbf{0.6131}}
        & 0.5564 / 0.6093
        & 0.5826 / 0.6067
        & \textbf{0.5875} / 0.6090 \\

        Mask Precision $\uparrow$
        & \underline{0.6577} / \underline{\textbf{0.7040}}
        & 0.6539 / 0.7015
        & 0.6726 / 0.6966
        & \textbf{0.6825} / 0.7021 \\

        Mask Recall $\uparrow$
        & \underline{0.7397} / \underline{\textbf{0.7810}}
        & 0.7263 / 0.7771
        & 0.7536 / 0.7753
        & \textbf{0.7578} / 0.7791 \\

        \addlinespace[4pt]

        Depth MAE (mm) $\downarrow$
        & 24.197 / 24.181
        & \underline{\textbf{23.909}} / \underline{\textbf{23.740}}
        & 24.227 / 24.366
        & 24.091 / 24.093 \\

        Depth Coverage $\uparrow$
        & \underline{0.7714} / \underline{\textbf{0.8060}}
        & 0.7574 / 0.8030
        & 0.7811 / 0.7997
        & \textbf{0.7864} / 0.8050 \\

        One-Way $L_1$ Chamfer (mm) $\downarrow$
        & \underline{14.894} / \underline{9.599}
        & 16.794 / 10.150
        & \textbf{12.947} / \textbf{8.317}
        & 14.793 / 10.104 \\

        \midrule

        \multicolumn{5}{@{}l}{\textit{Future Prediction}} \\
        \addlinespace[2pt]

        PSNR (dB) $\uparrow$
        & \underline{10.049} / \underline{\textbf{10.604}}
        & 10.027 / 10.551
        & 10.218 / 10.463
        & \textbf{10.308} / 10.590 \\

        SSIM $\uparrow$
        & 0.9778 / 0.9794
        & \underline{0.9780} / \underline{0.9796}
        & 0.9784 / 0.9793
        & \textbf{0.9788} / \textbf{0.9797} \\

        LPIPS $\downarrow$
        & 0.0282 / 0.0235
        & \underline{0.0279} / \underline{0.0230}
        & 0.0260 / 0.0237
        & \textbf{0.0252} / \textbf{0.0230} \\

        \addlinespace[4pt]

        Mask IoU $\uparrow$
        & \underline{0.4099} / 0.4887
        & 0.3993 / \underline{0.4975}
        & 0.4461 / 0.4794
        & \textbf{0.4502} / \textbf{0.5001} \\

        Mask Precision $\uparrow$
        & \underline{0.5366} / 0.6044
        & 0.5280 / \underline{0.6115}
        & 0.5624 / 0.5948
        & \textbf{0.5690} / \textbf{0.6132} \\

        Mask Recall $\uparrow$
        & \underline{0.6185} / 0.6977
        & 0.5978 / \underline{0.7028}
        & 0.6562 / 0.6914
        & \textbf{0.6569} / \textbf{0.7085} \\

        \addlinespace[4pt]

        Depth MAE (mm) $\downarrow$
        & \underline{\textbf{30.920}} / 33.765
        & 31.757 / \underline{\textbf{32.972}}
        & 32.988 / 34.494
        & 33.371 / 34.923 \\

        Depth Coverage $\uparrow$
        & \underline{0.6651} / 0.7331
        & 0.6330 / \underline{0.7360}
        & \textbf{0.6935} / 0.7205
        & 0.6852 / \textbf{0.7386} \\

        One-Way $L_1$ Chamfer (mm) $\downarrow$
        & \underline{25.457} / 17.429
        & 29.662 / \underline{17.354}
        & 25.383 / 18.758
        & \textbf{25.270} / \textbf{16.783} \\

        \bottomrule

    \end{tabular*}

\end{table*}

\subsubsection{Controller trajectory source}
We compare controller-point trajectories deterministically derived from robot joint states with those obtained through MV-TAP~\citep{koo2025mvtap} visual tracking, depth lifting, and PhysTwin-style controller filtering. On the same 26 robot-collected trajectories, robot-state-derived controllers enable successful reconstruction in all 26 cases, compared with only two cases for visually tracked controllers, corresponding to success rates of 100.0\% and 7.7\%, respectively. This comparison demonstrates the reliability benefit of robot-state-derived controllers under the tested conditions.

\subsubsection{Static geometry prior and object tracking methods}
We ablate the object tracking method and the pre-interaction geometry prior by evaluating CoTracker3~\citep{karaev2024cotracker3}, MVTracker~\citep{rajic2025mvtracker}, and MV-TAP~\citep{koo2025mvtap} with and without the prior, as shown in Table~\ref{tab:tracker_ablation}. CoTracker3 uses either mask-based or point-cloud-based initialization. The mask variant samples 5,000 points per view and follows PhysTwin's track-filtering procedure~\citep{phystwin}; the point-cloud variant, MVTracker, and MV-TAP each sample 1,024 points from the filtered first-frame object point cloud. CoTracker3 tracks projected queries independently per camera, while MV-TAP exploits multi-view visibility. MVTracker directly tracks 3D queries without depth lifting.

The geometry prior improves most metrics across all tracking configurations. Among methods using point-cloud initialization, MV-TAP generally performs best in both resimulation and future prediction. With the geometry prior, CoTracker3-m leads on most resimulation metrics, whereas MV-TAP leads on most future-prediction metrics.

On the robot-collected data, MV-TAP and MVTracker retain similar average numbers of tracking points per trajectory, at 615.96 and 625.46 respectively, while CoTracker3-m and CoTracker3-p retain 2,853.46 and 1,313.92. With the geometry prior, CoTracker3-m generally achieves better resimulation than CoTracker3-p, but worse future prediction. This trade-off mirrors the PhysTwin--AuRA comparison in Table~\ref{tab:reconstruction_quality}: PhysTwin retains an average of 1,493.76 points per trajectory, compared with AuRA's 615.96, and performs better on most resimulation metrics but worse on most future-prediction metrics. We hypothesize that denser tracking supervision helps fit observed motion but does not necessarily improve future prediction.

\subsubsection{Static versus first-frame Gaussians}
We compare static and first-frame Gaussian models on the same 13 unseen interactions using AuRA assets, keeping the simulated motion fixed. As shown in Table~\ref{tab:unseen_interaction_quality}, static Gaussians improve mask IoU from 0.3294 to 0.3861 and PSNR from 9.622 to 10.379~dB, slightly exceeding PhysTwin's 0.3849 and 10.309~dB, respectively. Mask recall and depth coverage both increase from 0.5821 to 0.6522, although depth MAE rises from 18.478 to 19.522~mm. These results show improved appearance and surface coverage with pre-interaction captures, accompanied by a modest increase in depth error.

%% file: sections/05_Discussion_and_Future_Work.tex
\section{Conclusion}

In this work, we present AuRA, an autonomous pipeline for interaction-data acquisition and physics-informed digital-twin reconstruction. AuRA combines execution-aware grasp selection, robot-state-derived controller trajectories, and pre-interaction geometry and appearance priors to reduce manual effort, avoid visual gripper tracking, and mitigate robot-induced occlusion. Across 13 deformable objects, AuRA reduces both human labor and collection time while achieving a higher end-to-end reconstruction yield than human-operated acquisition. Furthermore, compared with the human-collected PhysTwin baseline, AuRA achieves comparable reconstruction quality. Together, these results show that autonomous robot interaction can make real-to-sim asset acquisition more efficient and reliable.

%% file: sections/06_Acknowledgments.tex
\section*{ACKNOWLEDGMENT}

Jung, Chen, Lee and Huang are supported by DARPA HR001124S0029-AIQ-FP-019, National Science Foundation TRAILS Institute (2229885). Private support was provided by Open Philanthropy and Apple. The authors acknowledge the National Artificial Intelligence Research Resource (NAIRR) Pilot for contributing to this research result.
We thank Wenqi Marcus Wu for helping validate the human-operated data collection pipeline and Shengming Qi for annotating human involvement using videos recorded during data collection with a separate external camera.

%% file: example.bib
@String(CVPR  = {CVPR})

@String(ICCV  = {ICCV})

@String(ECCV  = {ECCV})

@String(NIPS  = {NeurIPS})

@string(ICRA = {ICRA})

@string(CoRL = {CoRL})

@string(RSS  = {RSS})

@string(ICML = {ICML})

@string(IROS = {IROS})

@inproceedings{kim2025pyroki,
  title     = {{PyRoki}: A Modular Toolkit for Robot Kinematic Optimization},
  author    = {Kim, Chung Min and Yi, Brent and Choi, Hongsuk and Ma, Yi and Goldberg, Ken and Kanazawa, Angjoo},
  booktitle = IROS,
  year      = {2025}
}

@inproceedings{MPM,
  title     = {The Material Point Method for Simulating Continuum Materials},
  author    = {Jiang, Chenfanfu and Schroeder, Craig and Teran, Joseph and Stomakhin, Alexey and Selle, Andrew},
  booktitle = {ACM SIGGRAPH 2016 Courses},
  year      = {2016}
}

@inproceedings{huang2026soma,
  title     = {{SoMA}: A Real-to-Sim Neural Simulator for Robotic Soft-Body Manipulation},
  author    = {Huang, Mu and Wang, Hui and Ren, Kerui and Xu, Linning and Zhou, Yunsong and Yu, Mulin and Dai, Bo and Pang, Jiangmiao},
  booktitle = ICML,
  year      = {2026}
}

@inproceedings{zhang2024particle,
  title     = {Particle-Grid Neural Dynamics for Learning Deformable Object Models from {RGB-D} Videos},
  author    = {Zhang, Kaifeng and Li, Baoyu and Hauser, Kris and Li, Yunzhu},
  booktitle = RSS,
  year      = {2025}
}

@article{chenhu2026empm,
  title     = {{EMPM}: Embodied {MPM} for Modeling and Simulation of Deformable Objects},
  author    = {Chen, Yunuo and Hu, Yafei and Sun, Lingfeng and Kusnur, Tushar and Herlant, Laura and Jiang, Chenfanfu},
  journal   = {IEEE Robotics and Automation Letters},
  volume    = {11},
  number    = {4},
  pages     = {4179--4186},
  year      = {2026},
  publisher = {IEEE}
}

@inproceedings{koo2025mvtap,
  title     = {{MV-TAP}: Tracking Any Point in Multi-View Videos},
  author    = {Koo, Jahyeok and Kim, In{\`e}s Hyeonsu and Kim, Mungyeom and Park, Junghyun and Park, Seohyeon and Kim, Jaeyeong and Yi, Jung and Cho, Seokju and Kim, Seungryong},
  booktitle = CVPR,
  year      = {2026}
}

@inproceedings{lee2026physhandi,
  title     = {{PhysHanDI}: Physics-Based Reconstruction of Hand-Deformable Object Interactions},
  author    = {Lee, Jihyun and Lee, Changmin and Kim, Donghwan and Kim, Tae-Kyun},
  booktitle = ICML,
  year      = {2026}
}

@InProceedings{karaev2024cotracker3,
    author    = {Nikita Karaev and Iurii Makarov and Jianyuan Wang and Natalia Neverova and Andrea Vedaldi and Christian Rupprecht},
    title     = {{CoTracker3}: Simpler and Better Point Tracking by Pseudo-Labelling Real Videos},
    booktitle   = ICCV,
    year      = {2025}
  }

@inproceedings{rajic2025mvtracker,
  title     = {Multi-View {3D} Point Tracking},
  author    = {Raji{\v{c}}, Frano and Xu, Haofei and Mihajlovic, Marko and Li, Siyuan and Demir, Irem and G{\"u}ndo{\u{g}}du, Emircan and Ke, Lei and Prokudin, Sergey and Pollefeys, Marc and Tang, Siyu},
  booktitle = ICCV,
  year      = {2025}
}

@article{ren2024grounded,
  title     = {Grounded {SAM}: Assembling Open-World Models for Diverse Visual Tasks},
  author    = {Ren, Tianhe and Liu, Shilong and Zeng, Ailing and Lin, Jing and Li, Kunchang and Cao, He and Chen, Jiayu and Huang, Xinyu and Chen, Yukang and Yan, Feng and Zeng, Zhaoyang and Zhang, Hao and Li, Feng and Yang, Jie and Li, Hongyang and Jiang, Qing and Zhang, Lei},
  journal   = {arXiv preprint arXiv:2401.14159},
  year      = {2024}
}

@inproceedings{gong2023arnold,
  title     = {{ARNOLD}: A Benchmark for Language-Grounded Task Learning with Continuous States in Realistic {3D} Scenes},
  author    = {Gong, Ran and Huang, Jiangyong and Zhao, Yizhou and Geng, Haoran and Gao, Xiaofeng and Wu, Qingyang and Ai, Wensi and Zhou, Ziheng and Terzopoulos, Demetri and Zhu, Song-Chun and others},
  booktitle = ICCV,
  year      = {2023}
}

@inproceedings{margolisyang2022rapid,
  title     = {Rapid Locomotion via Reinforcement Learning},
  author    = {Margolis, Gabriel and Yang, Ge and Paigwar, Kartik and Chen, Tao and Agrawal, Pulkit},
  booktitle = RSS,
  year      = {2022}
}

@article{Acar_2024,
  title     = {Visual-Policy Learning Through Multi-Camera View to Single-Camera View Knowledge Distillation for Robot Manipulation Tasks},
  author    = {Acar, Cihan and Binici, Kuluhan and Tekirda{\u{g}}, Alp and Wu, Yan},
  journal   = {IEEE Robotics and Automation Letters},
  volume    = {9},
  number    = {1},
  pages     = {691--698},
  year      = {2024},
  publisher = {IEEE}
}

@inproceedings{torne2024rialto,
  title     = {Reconciling Reality Through Simulation: A Real-to-Sim-to-Real Approach for Robust Manipulation},
  author    = {Torne, Marcel and Simeonov, Anthony and Li, Zechu and Chan, April and Chen, Tao and Gupta, Abhishek and Agrawal, Pulkit},
  booktitle = RSS,
  year      = {2024}
}

@inproceedings{wang2025vggt,
  title     = {{VGGT}: Visual Geometry Grounded Transformer},
  author    = {Wang, Jianyuan and Chen, Minghao and Karaev, Nikita and Vedaldi, Andrea and Rupprecht, Christian and Novotny, David},
  booktitle = CVPR,
  year      = {2025}
}

@article{kerbl3Dgaussians,
  title     = {{3D} {Gaussian} Splatting for Real-Time Radiance Field Rendering},
  author    = {Kerbl, Bernhard and Kopanas, Georgios and Leimk{\"u}hler, Thomas and Drettakis, George},
  journal   = {ACM Transactions on Graphics},
  volume    = {42},
  number    = {4},
  pages     = {139:1--139:14},
  year      = {2023},
  publisher = {ACM}
}

@inproceedings{phystwin,
  title     = {{PhysTwin}: Physics-Informed Reconstruction and Simulation of Deformable Objects from Videos},
  author    = {Jiang, Hanxiao and Hsu, Hao-Yu and Zhang, Kaifeng and Yu, Hsin-Ni and Wang, Shenlong and Li, Yunzhu},
  booktitle = ICCV,
  year      = {2025}
}

@inproceedings{Qiao2021Differentiable,
  title     = {Differentiable Simulation of Soft Multi-body Systems},
  author    = {Qiao, Yi-Ling and Liang, Junbang and Koltun, Vladlen and Lin, Ming C.},
  booktitle = NIPS,
  year      = {2021}
}

@inproceedings{shi2022robocraft,
  title     = {{RoboCraft}: Learning to See, Simulate, and Shape Elasto-Plastic Objects with Graph Networks},
  author    = {Shi, Haochen and Xu, Huazhe and Huang, Zhiao and Li, Yunzhu and Wu, Jiajun},
  booktitle = RSS,
  year      = {2022}
}

@inproceedings{sanchezgonzalez2020learning,
  title     = {Learning to Simulate Complex Physics with Graph Networks},
  author    = {Sanchez-Gonzalez, Alvaro and Godwin, Jonathan and Pfaff, Tobias and Ying, Rex and Leskovec, Jure and Battaglia, Peter W.},
  booktitle = ICML,
  year      = {2020}
}

@inproceedings{lin2022learning,
  title     = {Learning Visible Connectivity Dynamics for Cloth Smoothing},
  author    = {Lin, Xingyu and Wang, Yufei and Huang, Zixuan and Held, David},
  booktitle = CoRL,
  year      = {2021}
}

@inproceedings{zhong2024springgaus,
  title     = {Reconstruction and Simulation of Elastic Objects with Spring-Mass {3D} {Gaussians}},
  author    = {Zhong, Licheng and Yu, Hong-Xing and Wu, Jiajun and Li, Yunzhu},
  booktitle = ECCV,
  year      = {2024}
}

@inproceedings{mildenhall2020nerf,
  title     = {{NeRF}: Representing Scenes as Neural Radiance Fields for View Synthesis},
  author    = {Mildenhall, Ben and Srinivasan, Pratul P. and Tancik, Matthew and Barron, Jonathan T. and Ramamoorthi, Ravi and Ng, Ren},
  booktitle = ECCV,
  year      = {2020}
}

@article{fang2023anygrasp,
  title     = {{AnyGrasp}: Robust and Efficient Grasp Perception in Spatial and Temporal Domains},
  author    = {Fang, Hao-Shu and Wang, Chenxi and Fang, Hongjie and Gou, Minghao and Liu, Jirong and Yan, Hengxu and Liu, Wenhai and Xie, Yichen and Lu, Cewu},
  journal   = {IEEE Transactions on Robotics},
  volume    = {39},
  number    = {5},
  pages     = {3929--3945},
  year      = {2023},
  publisher = {IEEE}
}

@article{hu2026autonomous,
  title     = {Autonomous Robotic Exploration of Unknown Soft Objects},
  author    = {Hu, Junlei and Jones, Dominic and Loza Galindo, Gerardo and Huang, Shoudong and Valdastri, Pietro},
  journal   = {The International Journal of Robotics Research},
  year      = {2026},
  publisher = {SAGE Publications}
}

@inproceedings{Dasgupta2024ActNeRF,
  title     = {{ActNeRF}: Uncertainty-aware Active Learning of {NeRF}-based Object Models for Robot Manipulators using Visual and Re-orientation Actions},
  author    = {Dasgupta, Saptarshi and Gupta, Akshat and Tuli, Shreshth and Paul, Rohan},
  booktitle = IROS,
  year      = {2024}
}

@article{mazza2026activecrossmodalvisuotactileperception,
  title     = {Active Cross-Modal Visuo-Tactile Perception of Deformable Linear Objects},
  author    = {Mazza, Raffaele and Natale, Ciro and Falco, Pietro},
  journal   = {arXiv preprint arXiv:2601.13979},
  year      = {2026}
}

@article{Yang2025DPSI,
  title     = {Differentiable Physics-based System Identification for Robotic Manipulation of Elastoplastic Materials},
  author    = {Yang, Xintong and Ji, Ze and Lai, Yu-Kun},
  journal   = {The International Journal of Robotics Research},
  volume    = {44},
  number    = {13},
  pages     = {2126--2155},
  year      = {2025},
  publisher = {SAGE Publications}
}

@inproceedings{zhang2024adaptigraph,
  title     = {{AdaptiGraph}: Material-Adaptive Graph-Based Neural Dynamics for Robotic Manipulation},
  author    = {Zhang, Kaifeng and Li, Baoyu and Hauser, Kris and Li, Yunzhu},
  booktitle = RSS,
  year      = {2024}
}

@inproceedings{Ma2023Sim2Real2,
  title     = {{Sim2Real$^2$}: Actively Building Explicit Physics Model for Precise Articulated Object Manipulation},
  author    = {Ma, Liqian and Meng, Jiaojiao and Liu, Shuntao and Chen, Weihang and Xu, Jing and Chen, Rui},
  booktitle = ICRA,
  year      = {2023}
}

@inproceedings{lv2022sagci,
  title     = {{SAGCI-System}: Towards Sample-Efficient, Generalizable, Compositional, and Incremental Robot Learning},
  author    = {Lv, Jun and Yu, Qiaojun and Shao, Lin and Liu, Wenhai and Xu, Wenqiang and Lu, Cewu},
  booktitle = ICRA,
  year      = {2022}
}
